# RESPONSES TO REVIEWERS' 1 COMMENTS

Ref:
Title: Early Detection of Late Blight Tomato Disease using Histogram Oriented Gradient based Support Vector Machine
Corresponding Author: Dr. Muhammad Hameed Siddiqi
Authors: Yousef Alhwaiti, Ph.D.; Muhammad Ishaq, Ph.D.; Muhammad Hameed Siddiqi, Ph.D.; Muhammad Waqas, Ph.D.; Madallah Alruwaili, Ph.D.; Saad Alanazi, Ph.D.; Asfandyar Khan, Ph.D.; Faheem Khan Ph.D.;

**[Acknowledgment]**
We would like to express our deep gratitude to anonymous reviewers for their insightful comments on the paper. We believe that their constructive suggestions have significantly enriched the quality of the paper. We have tried our best to modify the paper according to their recommendations.

- Any new information added to the paper as a response to reviewers' comments is presented in blue color in the revised paper.

Reviewer #1:

Question 1. Mature and excellent title with a comprehensive abstract.

**[Answer 1]**
**The authors panel is much obliged of this encouragement. We thoroughly reviewed and refined all sections of this article with senior and junior relevant interdisciplinary researchers. This research is purely motivated by huge losses to low and medium economy of local farmers community. Food security and end hunger is included in the United Nations Sustainable Development goals.**

Question 2. Algorithms and their statistical analyses, controls, sampling mechanisms, and statistical reporting (e.g., metrics) are appropriate and well described.

**[Answer 2]**
. **This research uses Support Vector Machine (SVM) and Histogram Oriented Gradient (HOG), the best computational algorithms, to detect pathogens in tomato leaf disease (Zhang et al., 2018).**

*The proposed hybrid algorithm of SVM and HOG has significant potential for the early detection of late blight disease in tomato plants. The performance of the proposed model against decision tree and KNN algorithms and the results may assist in selecting the best algorithm for future applications.*

Question 3. Existing tables and/or figures are complete and acceptable for publication. There are no specific suggestions for improvements, removals, or additions to figures or tables.

**[Answer 3]**
.**We develop and enhance images and tables with relevant and latest tools and techniques. The data in it is authentic and complete in all respects.**

Question 4. The interpretation of results and study conclusions are fully supported by the data. This article does not need any improvement, tone-down, or expansion of the study in terms of the interpretations and conclusions

**[Answer 4]**
**Beside online available datasets we also collected and improved images from the field. We also get help from relevant plant pathologists for correct identification of early blight.**

Question 5. The authors clearly emphasized the strengths of their research and proposed methods.

[Answer 5]
.Sustainable agriculture and food security is our goal. Smart agriculture is the only way out to increase tomato production. Tomato is much vulnerable to diseases and the only solution is in-time detection.

Question 6. Mathematical and statistical analyses, controls, sampling mechanisms, or statistical reporting are up to the mark.

[Answer 6]
**Datasets without raw field images can give much improved results. So there is always a chance of further improvements in the hybrid strategy through image enhancement.**

Question 7. The objectives and rationale of the study are clearly stated.

[Answer 7]
**. HOG is specifically used for feature extraction. SVM latest variants in the literature possess excellent feature extraction and classification.**

Question 8. Please provide improved image dataset links. If already provided in additional files, then ignore this comment.

[Answer 8]
. The dataset can be found at **https://www.kaggle.com/datasets/mamtag/tomato-village**
**The field from whci we collected images can be found at**
**Link to the Field Map:**

https://goo.gl/maps/UggWPAetbU83M8LK6

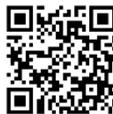

Figure 7: QR-Code to Map of Tomato Field

Question 9. Improve the clarity of the objectives and rationale of this research in the conclusion section.

[Answer 9] The following paragraph is added to the conclusion section.
. The true rational of this contribution is to achieve the sustainable development goal (food security) of the United Nations. Smart agriculture or AI enabled agriculture application can guarantee the end of hunger and help us to alleviate poverty.
This research achieved the following three objectives:
To enhance image dataset through inclusion of early affected late blight tomato leaves.
To propose HOG based SVM model for early detection of late blight tomatoes leaf disease.
To check the performance of the proposed model in terms of MSE, accuracy, precision and Recall as compare to Decision Tree and KNN.

Question 10. The objectives of the article are clear, but better organization of the manuscript is needed. Like why only late blight disease is used for identification, the author should include a solid reason.

[Answer 10]
. The pathogen Phytophthora infestans is the source of the fungal disease known as late blight that affects tomato plants. Fungi is a group heterotrophic plants that get their food from other resources (living or dead). It can result in severe crop losses and is a critical issue in warm, humid settings. Pakistan and

**specifically Peshawar valley is warm and humid. So it result in heavy losses to the farmer community. This research is purely motivated by huge losses to low and medium economy of local farmer's community.**

```
Question 11. The literature review includes some articles that are more than
ten years old. The author should consider including recent papers.
```

**[Answer 11]**
**.The origin of HOG is cited in reference No 4 and dated back to 2005. Type or encyclopedia of machine learning is cited in reference No 13. The rest of references are more recent quality contribution in this domain. The reference is up to date.**

```
Question 12. The manuscript does not mention limitations. The study has
limitations, e.g., there are more improved AI algorithms (Deep Learning) for
this purpose.
```

**[Answer 12]**
**.Feature extraction mechanism through HOG is state of the art. SVM latest variants performance can be compared with any CNN or RNN. More layers also result in high computational complexity. So for any real time farming practice HOG and SVM combination may help in the development of light application.**

```
Question 13. There lots of grammatical mistakes. Please improve the quality
of the paper by removing these mistakes.
```

**[Answer 13]**
**.Grammatical and sentence structure reviewed and potential corrections are incorporated in the article. There is no native speaker among us, so we extensively utilizes AI enabled Grammatical and sentence structure improvement tools.**

```
Question 14. The quality of the figures are very poor. Improve the quality
of the figures.
```

**[Answer 14]**
**.I reviewed recent articles in your journal, and I found that our figures and images are according to the standard journal requirements.**

# RESPONSES TO REVIEWERS' 2 COMMENTS

Ref:
Title: Early Detection of Late Blight Tomato Disease using Histogram Oriented Gradient based Support Vector Machine
Corresponding Author: Dr. Muhammad Hameed Siddiqi
Authors: Yousef Alhwaiti, Ph.D.; Muhammad Ishaq, Ph.D.; Muhammad Hameed Siddiqi, Ph.D.; Muhammad Waqas, Ph.D.; Madallah Alruwaili, Ph.D.; Saad Alanazi, Ph.D.; Asfandyar Khan, Ph.D.; Faheem Khan Ph.D.;

**[Acknowledgment]**
We would like to express our deep gratitude to anonymous reviewers for their insightful comments on the paper. We believe that their constructive suggestions have significantly enriched the quality of the paper. We have tried our best to modify the paper according to their recommendations.

- Any new information added to the paper as a response to reviewers' comments is presented in blue color in the revised paper.

Reviewer #2:

Question 1. Encounter an article with a real-time smart agriculture solution for better protected plants and sustainable food security.

**[Answer 1]**
**.Tomato is one of the major food ingredient. There are numerous value products associated with tomatoes. Disease identification can be carried out through Unmanned Ariel vehicle (UAV) equipped with high resolution camera or the farmers can use ordinary smart phones for image taking of infected plants.**

Question 2. As a future work, this research deserves funding to be implemented for real time plant protection even outside the Green House. Tomatoes are very vulnerable plants, and there are numerous diseases that can adversely affect their production.

**[Answer 2]**
**Identification of infected plants through UAV or ordinary smart phone camera is the first goal of this research effort. Infected plant ratio calculation inside the green house or in the ordinary agriculture land is another task. Farmer community should avoid the use of pesticides or other chemicals. FAO specifically recommend green solutions of all plant diseases. So how can we find out any green solution (without chemicals) for this disease using AI enabled applications?**

Question 3. The manuscript's structure, flow, or writing is acceptable for publication. Subheadings, shortening of text, reorganization of sections, or moving details from one section to another are not required. The article has improved manuscript structure and flow. But the current form also makes it understandable and read fluidly. The first paragraph after the keywords, the introduction section, literature review, method section, and results clearly fulfill the intended purpose.

**[Answer 3]**
**.We thoroughly reviewed recent relevant journals articles. The quality of article and optimal structure is due to the expertise of researchers in the interdisciplinary domain.**

Question 4. The article flow is fully synchronized. The structure of the article needs no revision.

**[Answer 4]**
**. Whenever we compose (article/patent) a well-organized and quality research contribution, then the flow of words is always synchronized.**

Question 5. The manuscript proposes the application of HOG-based SVM to identify tomato blight in its early stages. It also provides recommendations for improving the efficiency of the proposed technique.

**[Answer 5]**
**.Plant leaves first indicate the symptoms of the infection. So thousands of leaves images where added from ordinary fields for better trained disease detection application.**

Question 6. The quality of the figures is not bad. The author also includes the work of the computer vision libraries to improve image quality for better feature extraction.

**[Answer 6]**
**.Computer vision python libraries were used to prepare international standard image dataset pictures. Proper mapping was conducted of the field pictures with the online available relevant tomato dataset.**

Question 7. How to use a hybrid strategy in the detection of late blight disease The experimental results section is impressive.

**[Answer 7]**
**.In one type of hybridization we make changes to the respective code section of the algorithm in any library. The efficiency or the main theme of two algorithms were merged through coding. In another common type of hybrid strategy we order the steps of execution sequentially (most probably) of two or more algorithms.**

Question 8. AI-enabled smart disease identification is the only solution for futuristic agriculture. A comparison of HOG-based ensemble classifiers may further improve the article standard. It is advisable to follow the journal approach.

**[Answer 8]**
**.We added or amend many sections as per reviewer's comments. Yes the classifier level HOG-based ensemble improve accuracy, reduce overfitting and noise. Many interdisciplinary researchers are unable to interpret the comparison of HOG based ensemble classifiers.**

Question 9. It is advisable to include unmanned aerial vehicle (UAV)-taken images from the field.

**[Answer 9]**
**.UAV with high resolution cameras are easily available in the market. In a similar project for weed detection we requested the local police department to provide us UAV.**

Question 10. Make the article understandable for agriculture-side experts by explaining domain-specific terminologies.

**[Answer 10]**

. Much needed improvement were incorporated to make it understandable for plant protection experts. Even we added botanical and plant pathology terminologies. We are also thankful to the botanists and relevant plant pathologists for their help and contribution.

```
Question 11. The introduction must contain a paragraph introducing the
various sections of the article.
```

**[Answer 11]**
.See the following section for reference.

## Paper Organization

This paper is structured into seven sections, beginning with an introductory section that discusses the research topic of late blight disease detection and classification in tomato plants using machine learning techniques. It covers the research background and the significance of the study.

```
Question 12. Did the author create these figures, or were they adapted?
```

**[Answer 12]**
.We use various programming and graphical tools for image/figure creation, editing and improvement. Despite of extensive efforts we never achieve R studio standard figures.

```
Question 13. The literature review needs to be updated and consider papers
that are 3-5 years old. Why did the author not consider the IDLE tools that
are powered by Google or even a more recent Anaconda suite of compilers?
```

**[Answer 13]**
. The origin of HOG is cited in reference No 4 and dated back to 2005. Type or encyclopedia of machine learning is cited in reference No 13. The rest of references are more recent quality contribution in this domain. The reference is up to date.
We specifically used Google Colab and Anaconda based Jupyter note book (IDLE).

```
Question 14. The author must not use the 1st person singular (e.g., I have).
```

**[Answer 14]**
.Necessary corrections made. Unnecessary first person usage is removed from the article.

```
Question 15. What were the criteria for the selection of assessment tools
like algorithms and programming platforms?
```

**[Answer 15]**
.HOG-based SVM is improved and state of the art in performance. State of the art algorithm mean that we have less computational complexity and help us to develop light application for real time issue. Python is extensively with feature rich libraries and generic or platform independent nature.

```
Question 16. The author claims to propose an early detection method. However,
research is basically supported by the application of a hybrid methodology.
```

**[Answer 16]**
.Early detection mean to identify the tomato blight in the leaves. The disease infection is spread to other plant parts, more specifically to tomato fruit. Explanation and justification of hybrid strategy is already mentioned in the response 7.

```
Question 17. The manuscript does not mention limitations. The study has
limitations, as seen in the previous paragraph, among others.
```

**[Answer 17]**
**.A comprehensive explanation is mentioned in the response 7. Up to our findings the proposed method enable us to develop light application that can run on handheld devices. In Pakistan ordinary farmers have basic access to internet and smart phones.**

# RESPONSES TO REVIEWERS' 3 COMMENTS

Ref:
Title: Early Detection of Late Blight Tomato Disease using Histogram Oriented Gradient based Support Vector Machine
Corresponding Author: Dr. Muhammad Hameed Siddiqi
Authors: Yousef Alhwaiti, Ph.D.; Muhammad Ishaq, Ph.D.; Muhammad Hameed Siddiqi, Ph.D.; Muhammad Waqas, Ph.D.; Madallah Alruwaili, Ph.D.; Saad Alanazi, Ph.D.; Asfandyar Khan, Ph.D.; Faheem Khan Ph.D.;

**[Acknowledgment]**
We would like to express our deep gratitude to anonymous reviewers for their insightful comments on the paper. We believe that their constructive suggestions have significantly enriched the quality of the paper. We have tried our best to modify the paper according to their recommendations.

- Any new information added to the paper as a response to reviewers' comments is presented in blue color in the revised paper.

Reviewer #3:

Question 1. The application, theory, method, and study were reported with sufficient detail to allow for their replicability and/or reproducibility.

**[Answer 1]**
**.Research reusability is our prime target. The same kind of efforts is possible in cases of pests, weeds or insect control in any scope and capacity. It can be extended to other food producing plants. The discovery of first organic compound (Amonia) in the laboratory and the advent in smart agriculture revolutionize the agriculture production and associated industry. A new discussion section has been added to the article.**

Question 2. The article should include: Which specific SVM variant is used? and why?

**[Answer 2]**
**.Both twin and smooth SVM are equally efficient and effective. Twin is specifically used in cases when we are dealing with handheld devices and related datasets. Smooth SVM is good for noisy datasets. Twin is computationally expensive than standard SVM. Smooth accuracy is much lower than the standard SVM. So we prefer standard SVM.**

Question 3. A good article should also have a novel and improved methodology. The article possesses a reasonable and novel methodology with improved application of computer vision for image improvement.

**[Answer 3]**
**. International standard methodology of IBM for data science is more robust and can be followed by all researchers. We are committed to follow all standardized methodologies.**

Question 4. Find my comments below. As a reviewer, my aim is to provide feedback to authors. Acceptance and rejection of manuscripts, is the authority of the editor. Below are my comments for your guidance.

**[Answer 4]**
**.Respected Reviewers, We are thankful for useful comments and feedback. This document is the point by point response of all reviewers' comments. The Manuscript is updated accordingly.**

```
Question 5. The abstract is attractive; it has motivation that clearly
focuses on the importance of this research for sustainable agriculture. I
appreciate this extensive multidisciplinary effort for plant protection.
Local farmers in the community take pictures of their tomatoes and
immediately identify tomato blight disease. They can discover the type and
scale of disease for an effective preventive strategy.
```

[Answer 5]
.**Good comprehension of the main theme of the manuscript. AI can help us to alleviate poverty and to promote sustainable agriculture. Food security may end hunger and improve agriculture based economies specifically in the third world.**

```
Question 6. Nothing was missing, even though the author satisfactorily
explained agriculture-related terms and concepts.
```

[Answer 6]
.**Some terms are purely botanical and relevant to plant protection and plant pathology. The science of pathogens is called pathology. A pathogen is any living microbes, fungus, protozoans, algae or even angiosperm or gymnosperms that rely on other living creatures for food or shelter. Pathogen get food from host and severely damage it. In our case a fungus (pathogen) damage the tomatoes (host).**

```
Question 7. The result, discussion, and implications sections are impressive
and clear, with better visualization.
```

[Answer 7]
.**Our effort is a roadmap for other identical smart plant protection systems. For example if we detect the disease in early stage then the damage to tomatoes will be lower. The farmers may control it through the use of low quantity pesticides. WHO strictly ban some pesticides and encourage to abandon its uses.**

```
Question 8. Beside the terminology explanation, the introduction extended
the research motivation and research gap. The authors writes up; it seems
that he has concentrated on manuscript writing. The citations are correct,
and nothing is missing. All the citations are as per the standard of the
manuscript.
```

[Answer 8]
.**Several meetings were conducted with agriculture researchers for relevant terms and more refined research. We are non-natives' speakers, so we try to use small sentences. It is really a tedious task to find quality relevant references.**

```
Question 9. The author should follow proper headings and subheadings,
specifically in the literature review section. This article has organized
the literature, and previous studies have been properly acknowledged.
```

[Answer 9]
**We are indebted of the positive feedback and comments. The funding is acknowledged. Even plant pathology references were cited for proper multidisciplinary future impact.**
.
```
Question 10. The objective and problem statement are clear. The article
should follow the standard journal paper template.
```

[Answer 10]
.**The consist of objectives and detailed problem statement. In one research approach we raise an issues and then resolve it. Some research problems stem from the industry or society and call it real time issues. Our stated research issues and its solution belong to both types of research. Some authors are farmers and they were anxious to counter the huge damage to their tomato plants.**

```
Question 11. I also recommend grammar and spelling revision through native
reviewer or AI Generative tools for grammar and proper sentence structure.
```

**[Answer 11]**

.

**.Grammatical and sentence structure reviewed and potential corrections are incorporated in the article. There is no native speaker, so we extensively utilizes AI enabled Grammatical and sentence structure improvement tools.**

```
Question 12. Overall, I think the author should follow some top relevant
journal research papers and update the references section.
```

**[Answer 12]**
**. Already incorporated. The origin of HOG is cited in reference No 4 and dated back to 2005. Type or encyclopedia of machine learning is cited in reference No 13. The rest of references are more recent quality contribution in this domain. The reference is up to date.**